\documentclass[a4paper,twoside]{article}

\usepackage{epsfig}
\usepackage{subcaption}
\usepackage{calc}
\usepackage{amssymb}
\usepackage{amstext}
\usepackage{amsmath}
\usepackage{amsthm}
\usepackage{multicol}
\usepackage{pslatex}
\usepackage{apalike}
\usepackage{algorithm2e}
\usepackage{pifont}
\usepackage[bottom]{footmisc}
\usepackage{SCITEPRESS}     
\usepackage[inkscapelatex=false]{svg}
\begin{document}

\title{BEVMOSNet: Multimodal Fusion for BEV Moving Object
Segmentation}
\author{\authorname{
 Hiep Truong Cong\sup{1,3}, 
 Ajay Kumar Sigatapu\sup{1}, 
 Arindam Das\sup{2,3}, 
 Yashwanth Sharma\sup{2},
 Venkatesh Satagopan\sup{2},
 Ganesh Sistu\sup{3,4}
 and Ciar\'an Eising\sup{3}}
 \affiliation{\sup{1}DSW, Valeo Kronach, Germany, \sup{2}DSW, Valeo India}
 \affiliation{\sup{3}University of Limerick, Ireland and \sup{4}Valeo Vision Systems, Ireland}
 \email{firstname.lastname@\{valeo.com, ul.ie\}}
 }


\abstract{
Accurate motion understanding of the dynamic objects within the scene in bird's-eye-view (BEV) is critical to ensure a reliable obstacle avoidance system and smooth path planning for autonomous vehicles. However, this task has received relatively limited exploration when compared to object detection and segmentation with only a few recent vision-based approaches presenting preliminary findings that significantly deteriorate in low-light, nighttime, and adverse weather conditions such as rain. Conversely, LiDAR and radar sensors remain almost unaffected in these scenarios, and radar provides key velocity information of the objects. Therefore, we introduce BEVMOSNet, to our knowledge, the first end-to-end multimodal fusion leveraging cameras, LiDAR, and radar to precisely predict the moving objects in BEV. In addition, we perform a deeper analysis to find out the optimal strategy for deformable cross-attention-guided sensor fusion for cross-sensor knowledge sharing in BEV. While evaluating BEVMOSNet on the nuScenes dataset, we show an overall improvement in IoU score of 36.59\% compared to the vision-based unimodal baseline BEV-MoSeg \cite{sigatapu2023bev}, and 2.35\% compared to the multimodel SimpleBEV \cite{harley2022simple}, extended for the motion segmentation task, establishing this method as the state-of-the-art in BEV motion segmentation.}

\onecolumn \maketitle \normalsize \setcounter{footnote}{0} \vfill

\section{\uppercase{Introduction}}
\label{sec:introduction}

Recent research in accurate modeling of dynamic obstacles \cite{das2024fisheye} has stimulated rapid progress in achieving autonomous navigation in intricate environments, ensuring effective collision avoidance. Key components for safe and efficient autonomous driving include comprehending the movements of nearby objects and planning the vehicle's trajectory based on their anticipated future states. In recent times, the realm of autonomous driving has experienced notable progress, with leading car manufacturers integrating multiple sensor technologies \cite{xu2017learning,li2018multispectral,dasgupta2022spatio} to enhance the reliability of their autonomous systems.

\begin{figure}[!ht]
    \captionsetup{singlelinecheck=false, font=small, belowskip=-6pt}
    \centering
    \includegraphics[width=\columnwidth]{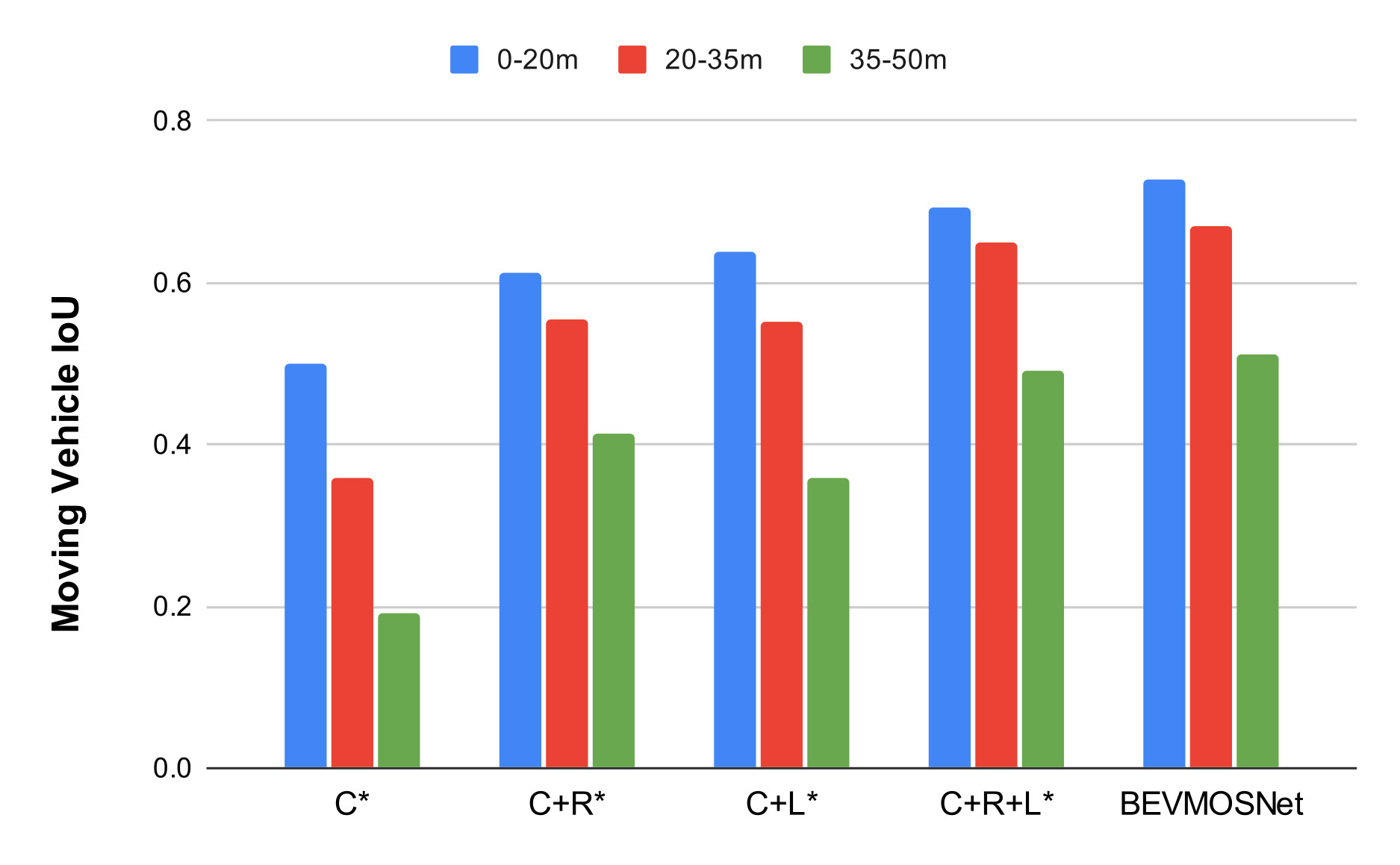}
    \vspace{-4mm}
    \caption{We propose BEVMOSNet for motion understanding within the scene. We demonstrate that our multi-modal fusion encompassing 6-cameras (C), LiDAR (L), and radar (R) yields better IoU across all distance ranges when compared to camera-only and other multimodal models. * denotes the SimpleBEV baseline model extended for the motion segmentation task.}
    \label{fig:iou_vs_distance}
\end{figure}

Rich semantic information in the image pixels has motivated the research community to pursue perception in Bird's-Eye-View (BEV) space \cite{roddick2020predicting,philion2020lift}. Despite the low cost and several other advantages, cameras are prone to failure in low illumination (low light, low contrast) and adverse weather conditions \cite{das2020tiledsoilingnet}. We aim to alleviate the failure in low illumination conditions by using LiDAR technology (another rich semantic 3D information-providing sensor).

Despite all the advantages of the camera and LiDAR sensor, they are prone to failure in adverse weather conditions \cite{godfrey2023evaluation}. We aim to address this key shortcoming by integrating another cost-effective sensor, such as radar \cite{dong2020probabilistic}. It is highly reliable in adverse weather conditions and supports long-range perception. In addition, it holds key velocity information that catalyzes the detection of moving vehicles. However, radar has limitations such as the sparsity of the points in a single frame \cite{lippke2023exploiting} when compared to LiDAR in the nuScenes dataset \cite{nuscenes}. We aim to make use of the complimentary features from all three sensors. Figure \ref{fig:iou_vs_distance} shows the superiority in terms of distance-based IoU metrics of all three sensors when fused together over camera only, camera + radar, and camera + LiDAR fusion proposals, respectively on the nuScenes dataset.

Recent works such as \cite{man2023bev,liang2022bevfusion} focus on producing an accurate BEV semantic representation of the surrounding 3D space using a fusion of multiple sensors like multi-view cameras, radar, and LiDAR, as the BEV coordinates acts as a common ground for representing the sensor-agnostic information.  Methods similar to \cite{chen2023futr3d} are prone to errors, as the imaging modality requires explicit depth estimation, the reason being the transformation process is quite complex, and any error in this process will have an impact on the subsequent fusion.

This paper presents evidence highlighting the significant impact of automotive sensors beyond cameras for the task at hand. We essentially paid more focus on how to intelligently integrate radar and LiDAR with camera sensors to propose a robust automotive multisensor perception stack. Our first approach is to project the sensor agnostic rich semantic features into a common reference like BEV and combine all of them by simple element-wise fusion as concatenation \cite{harley2022simple}. However, the fused features suffer from misalignment due to a significant domain gap between the modalities. For example, the camera has rich semantic features but an inaccurate spatial representation due to an ambiguous transformation process. On the contrary, radar has weak semantic cues but an accurate spatial position. We use the multimodal deformable cross-attention (\textbf{MDCA}) to share the cross-modal knowledge in BEV space.

The main contributions of this paper are as follows.
\begin{itemize}
  \item We present a novel multisensor deep network \textit{BEVMOSNet}, designed specifically for precise motion understanding in a bird's-eye-view. The proposed network combines multi-view cameras, LiDAR, and radar, representing the first known endeavor of its kind.
  \item Deformable cross attention (DCA) guided design of a sensor fusion module encompassing three modalities to democratize knowledge from individual sensors to cross modalities in BEV.
  \item Implementation of a single-stage end-to-end trainable network establishing the first state-of-the-art results on the nuScenes dataset, at the same time improving respective state-of-the-art performance for the camera-only proposal. 
  \item We perform thorough ablation studies considering a range of backbones, network components, and diverse feature fusion techniques.
\end{itemize}

\section{\uppercase{Related work}}
\label{sec:related_work}

Moving object segmentation is the task of understanding the dynamic properties in a scene. It includes the detection of moving objects and segmenting them from background or static components. Traditional computer vision techniques such as optical flow have been proposed to estimate the movement at the pixel level in a sequence of images. The limitation of optical flow is that it is only able to estimate relative pixel displacements between two consecutive frames and cannot distinguish dynamic and static components. Many other methods have attempted to overcome this limitation by estimating background motion \cite{Wehrwein2017} and using RGB-D data \cite{Menze2015}.

\begin{figure*}[ht]
    \captionsetup{singlelinecheck=false, font=small, belowskip=-6pt}
    \centering
    \includegraphics[width=\textwidth]{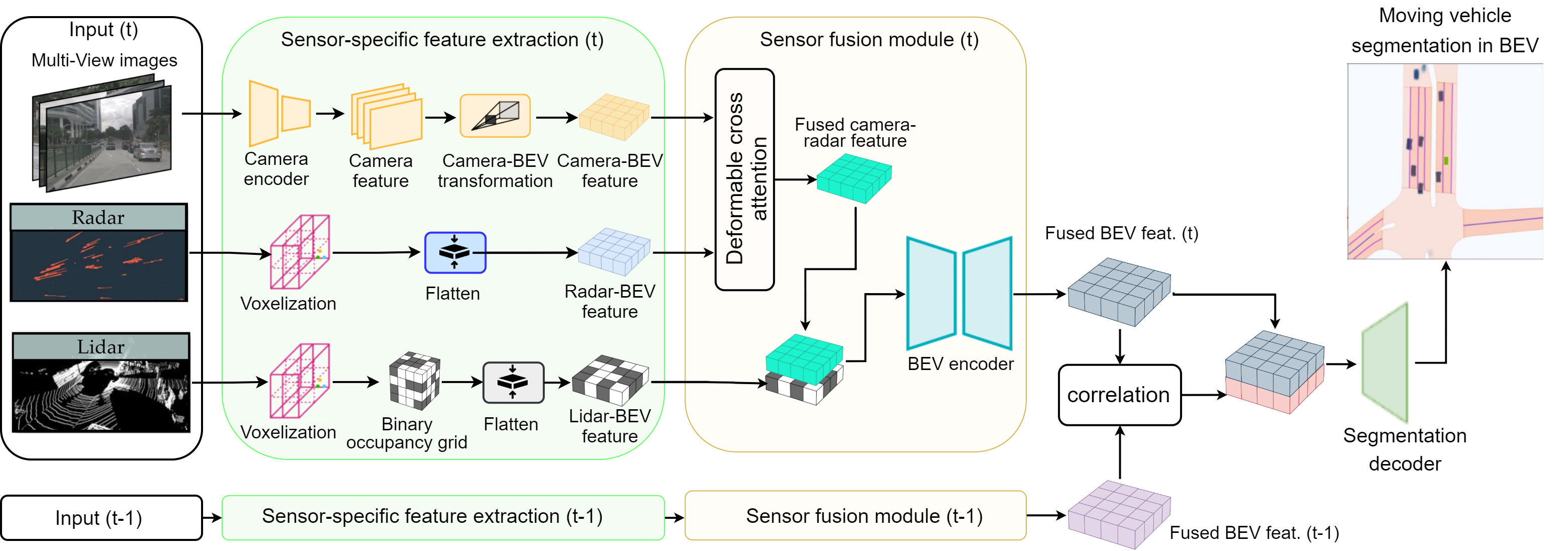}
    \caption{BEVMOSNet extracts features from camera, radar, and LiDAR input and transforms them into BEV, where they are fused together by a sensor fusion module. Consequently, a correlation block is applied to the fused BEV feature maps from current and previous frames to extract motion cues, which are then combined with the current fused BEV feature map as input for the segmentation decoder.}
    \label{fig:Architecture}
\end{figure*}

With the emergence of efficient deep learning networks that boost the performance of perception tasks, many researchers have been focusing on leveraging learning-based methods for motion segmentation. In \cite{Patil2020} an end-to-end, multi-frame multi-scale encoder-decoder adversarial learning network is proposed for
moving object segmentation. \cite{Fragkiadaki2015} uses a CNN with a dual-pathway architecture operating on both RGB images and optical flow to estimate moving objects. InstanceMotSeg \cite{mohamed2020instancemotseg} employs the flow field as an extra source of information, guiding a deep learning model to understand object motion at the instance level. In such a multimodal setup, ensuring minimal modality imbalance \cite{das2023revisiting} is always challenging with automotive sensors.

In the last few years, many large-scale multi-modal datasets for autonomous driving have been released, e.g., NuScenes \cite{nuscenes}, Waymo \cite{Sun_2020_CVPR}. These datasets provide 3D data, such as LiDAR, radar, and surround-view camera data. This facilitated the study of 3D perception. Many researchers targeted moving object segmentation in 3D LiDAR data. For instance, \cite{chen2021ral} segmented LiDAR points corresponding to moving objects using range images generated from point clouds. Instead of using range images as a secondary input, \cite{mohapatra2022limoseg} used only point cloud sequences to segment LiDAR points on moving objects and achieved real-time performance. Further studies in LiDAR moving object segmentation are InsMOS \cite{wang2023arxiv}, MotionBEV \cite{motionbev2023} and MambaMOS \cite{zeng2024mambamos}.

Recently, many studies have been investigating the environmental perception in Bird's Eye View (BEV) space. Because BEV is a natural representation of 3D space with the vertical dimension compressed, perception models that operate in BEV are not only more efficient, but also have competitive performance. One of the pioneering works is BEV-MODNet \cite{Hazem2021} which segments moving vehicles in BEV just using a monocular front camera. Similar to \cite{Fragkiadaki2015} and \cite{mohamed2020instancemotseg}, this work also leverages optical flow as the second input to predict moving vehicles in BEV. Many other studies utilize the multi-view camera data in large-scale datasets, such as Lift-Splat-Shoot (LSS) \cite{philion2020lift}, which proposed a learning-based method to perform semantic segmentation in BEV using multi-view camera data. Following LSS, BEV-MoSeg \cite{sigatapu2023bev} added a correlation layer on top of two BEV feature maps from two consecutive frames to predict feature correspondence in BEV space before utilizing another convolution layer to predict movements of moving vehicles in the current frame. Beyond moving object segmentation in BEV, some other research tackles the dynamic perception problem in the form of motion prediction. Fiery \cite{fiery2021}, PowerBev \cite{powerbev2023} and TBP-Former \cite{Fang_2023_CVPR} utilize only surround views images to target future instance segmentation and motion at the same time.

Recent studies have expanded the perception tasks to radar data, as it contains object velocity, which is valuable information for dynamic perception. RaTrack \cite{pan2023moving} proposes a network for moving object detection and tracking only based on radar. RadarMOSEVE \cite{pang2024radarmoseve} proposes a Spatial-Temporal Transformer Network for moving object segmentation and ego-velocity estimation. Radar Velocity Transformer \cite{RadarVelTrans2023} targets moving object segmentation tasks using only a single-scan radar point cloud. This work is also extended for moving instance segmentation tasks as in \cite{RadarInstanceTrans2024}. To our knowledge, there are no previous works that tackle the MOS task in BEV space by leveraging the multisensor data. To address this, we introduce BEVMOSNet, a multimodal deep learning model for precise motion understanding in the BEV space.

\section{\uppercase{Proposed Approach}}

In this section, we describe our overall architecture and the different fusion methodologies we employed for motion segmentation in BEV space. Our proposed method consists of a multimodal feature extraction module, followed by a sensor fusion module that includes a multi-headed deformable cross-attention strategy. Additionally, a correlation module is used to extract temporal features across multiple frames in BEV, and a segmentation decoder is employed to precisely segment objects from the correlated features. Our proposed model utilizes camera, radar, and LiDAR sensor data provided in the nuScenes dataset. 

\subsection{Sensor-specific feature extraction} \label{feature_extraction}
The module takes the raw camera, LiDAR and radar data as inputs and extracts three sets of feature maps in BEV corresponding to each modality. In this work, we follow the multi-stream setup in SimpleBEV \cite{harley2022simple} to extract multimodal features, which consists of a CNN-based camera feature extractor, a LiDAR, and a radar voxelization module respectively.

\textbf{Camera feature extractor.} We reuse the camera feature extractor from \cite{harley2022simple}, which consists of an image encoder and a 2D-3D lifting module. The input RGB images, shaped $3 \times H \times W$, are fed into a ResNet-101 \cite{he2016deep} backbone. The output from layer 3 is upsampled and concatenated with the layer 2 output before being processed by two additional CNN blocks with instance normalization and ReLU activation. A final convolution layer reduces the number of channels to create image feature maps with shape $C \times H/8 \times W/8$. These 2D feature maps are then transformed into BEV space using the lifting module from \cite{harley2022simple}. Where each 3D voxel “pulls” a feature from the 2D map, by projection and subpixel sampling. This results in a 3D feature volume with shape $C \times X \times Y \times Z$. Finally, this volume is rearranged to yield an image BEV feature map with shape $(C \times Y) \times X \times Z$.

\textbf{LiDAR feature extractor.} We voxelize the input LiDAR point cloud to create a binary occupancy grid with the shape of $Y \times Z \times X$. In this work, we aim to focus on the sensor fusion and keep the point cloud features as simple as possible. We only leverage the LiDAR points to provide our model with the information about object locations.

\textbf{Radar feature extractor.} In our baseline architecture, we rasterize the radar point clouds to create a radar BEV feature map. In the nuScenes dataset, each radar point has 18 attributes; the first 3 positions are point locations, and the remainder consists of velocity, compensated velocity, and other built-in pre-processing information. We use the first three attributes for rasterizing the radar point cloud, and we keep the other 15 attributes as radar features. For the MOS task, we focus on extracting dynamic information about moving objects from velocity attributes rather than relying on the radar point location information.

\subsection{Sensor fusion module} \label{SensorFusionModule} This module introduces the sensor-specific features from the unimodal encoders. We present several fusion strategies to determine the optimal configuration among the modalities.

\textbf{Concatenation: }
We start with the simple concatenation fusion method as proposed in SimpleBEV \cite{harley2022simple}. This serves as the baseline for our proposed strategies in the following sections. SimpleBEV \cite{harley2022simple} follows a compression of BEV features to reduce the feature dimension of the unified BEV feature map. Hence, we followed the same techniques for all sensor fusion approaches. 

\begin{figure*}[ht]
    \captionsetup{singlelinecheck=false, font=small, belowskip=-6pt}
    \centering
    \includegraphics[width=0.7\textwidth]{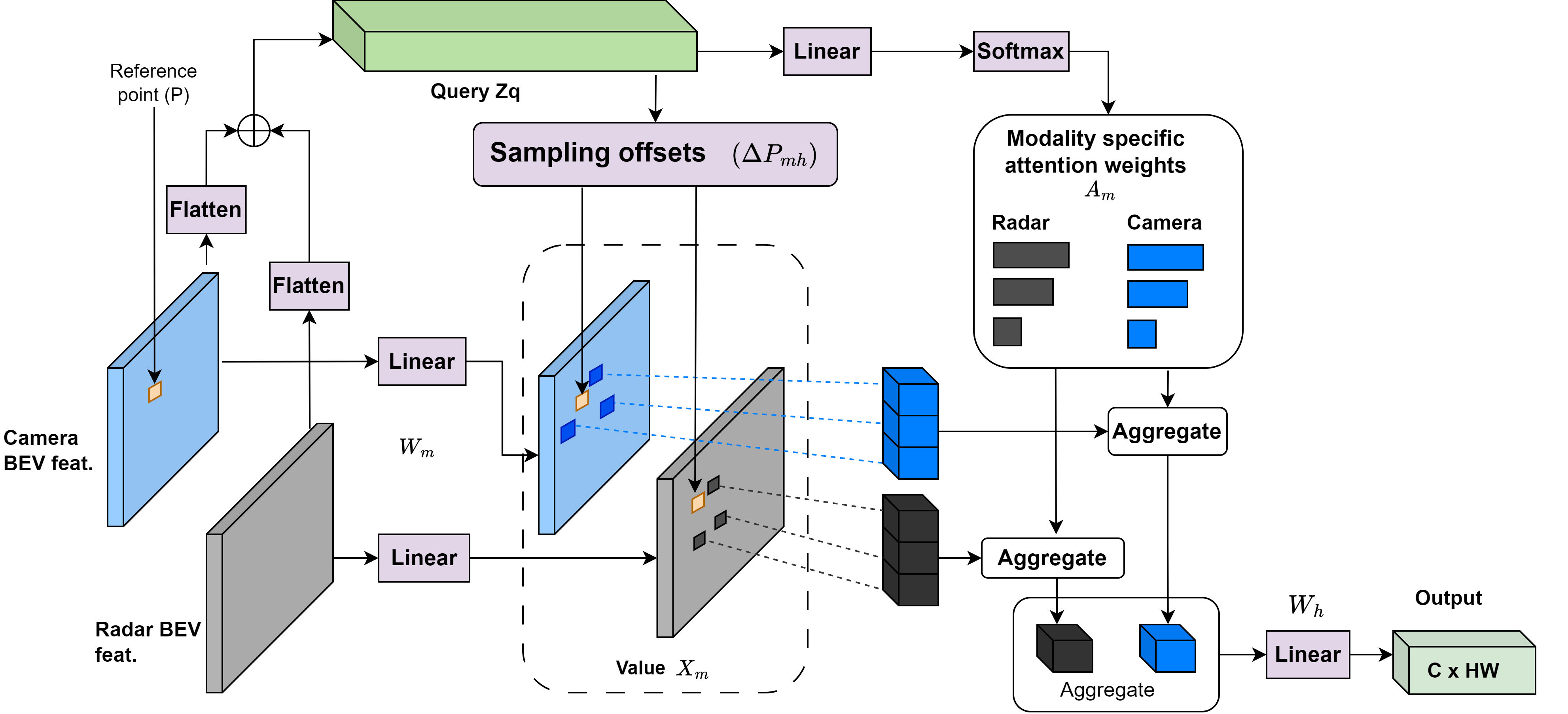}
    \caption{Multimodal deformable cross attention (MDCA) extracts complementary features from camera and radar sensors individually by separately applying attention weights $\mathbf{A}_{m}$ and learnable sampling offsets $\mathbf{\triangle P}_{m,h}$ in every attention head. $\oplus$ denotes concatenation.
    }
    \label{fig:def_attn}
\end{figure*}

\textbf{Multi-modal Deformable Cross-Attention (MDCA):} Cross-attention has proven to yield effective results in multimodal fusion applications. However, the computation cost is quadratic to the length of the input vector $O(N^2)$, where $N = X\times Z$ and $X, Z$ denote the height and width of the Bird's Eye View feature map. Taking into account the computational cost, we use the deformable multimodal cross attention \cite{kim2023crn} in the present task.

We apply deformable multi-modal cross attention shown in Figure \ref{fig:def_attn}. This mechanism selectively attends to a small set of keys sampled around a reference point in the spatial dimension for each query. This allows us to effectively identify and track moving vehicles in the neighboring pixels.

Given the sensor-agnostic BEV feature maps, we flatten them to obtain $\mathbf{\textit{I}} \in \mathbb{R}^{{C_r} \times XZ}$ and $\mathbf{\textit{R}} \in \mathbb{R}^{{C_c} \times XZ}$, where subscripts $r$ and $c$ denote radar and camera, respectively. $\mathbf{Z_q} \in \mathbb{R}^{XZ \times C}$ is the result of the linear projection of $\mathbf{\textit{I}}^\top \oplus \mathbf{\textit{R}}^\top$. We aim to enrich the BEV feature map using multi-head multi-modal cross-attention (MDCA) as described:

\begin{multline}\label{eq:1}
    \mathbf{MDCA}(\mathbf{Z}_q,\mathbf{P}_q,\mathbf{X}_m)= \\
    \sum_{h=1}^{H}\Bigg[\sum_{m=1}^{M}\mathbf{A}_{m,h}\mathbf{X}_m(\mathbf{P}+\Delta\mathbf{P}_{m,h})\mathbf{W}_m^\top\Bigg]\mathbf{W}_h^\top
\end{multline}

Where, $\mathbf{A}_{m,h}=\textit{softmax}(\mathbf{W}_{m,q}\mathbf{Z}_q)$  and $\Delta\mathbf{P}_m=\mathbf{W}_{m,q}^{\prime}\mathbf{Z}_q$ 
are obtained by linear projection over the queries.

${\mathbf{P}} \in \mathbb{R}^{X \times Y \times 2}$ is the reference point matrix, ${\boldsymbol{\Delta P}}_{m,h} \in \mathbb{R}^{X \times Y \times 2}$ is the offset matrix, and $A_{m,h} \in \mathbb{R}^{XY \times K}$ is the attention weight matrix of the $h$-th attention head, where $A_{m,h} X_m(P+\boldsymbol{\Delta P}_{m,h}) \in \mathbb{R}^{XY \times C}$. ${\mathbf{W}_h} \in \mathbb{R}^{C_v \times C}$ and ${\mathbf{W}_m} \in \mathbb{R}^{C \times C_v}$, here $h$ and $m$ index the attention head and modality, respectively.

\subsection{BEV encoder}
As mentioned in BEVFusion \cite{liu2022bevfusion}, despite deformable cross attention being applied for camera and radar and all sensor-specific BEV feature maps being in the same space, there are still local misalignments between them. The camera features in BEV are not accurately located due to errors in the view transformation. Radar and LiDAR BEV feature maps are also not aligned perfectly because they have different sparsity, and radar data is noisy. To this end, we apply a BEV encoder block, which is based on ResNet18 to compensate for the misalignments.

\subsection{Correlation for detecting motion}
We aim to extract the motion cues from the scene by analyzing a pair of consecutive temporal frames similar to BEV-MoSeg \cite{sigatapu2023bev}. Recent pixel-based optical flow  \cite{dosovitskiy2015flownet}, initially processes each image independently using convolutional neural networks (CNNs) to extract feature representations. Subsequently, akin to traditional computer vision methods that compare features from image patches,  the network correlates these learned representations at a higher level to identify relationships between the two images.

We take pixel-based flow estimation a step further by applying correlation layers to expand it into a higher-dimensional space representing a bird's-eye view (BEV), these correlation layers allow the network to compare sub-regions  from $f_1$ with all other sub-regions in $f_2$. This enables the network to capture more complex relationships between the two images. The "correlation" between image patches, centered at $x_1$ in the first feature map and $x_2$ in the second feature map, was defined as in \cite{dosovitskiy2015flownet}:
\begin{equation}
\mathbf{c}(x_1, x_2) = \sum_{o \in [-k,k] \times [-k,k]} \langle \mathbf{f_1}(x_1+o), \mathbf{f_2}(x_2+o) \rangle
\end{equation}
Here, $K := 2k + 1$ represented the size of a square kernel. In our experiments, we have used $k = 3$.  While the presented equation resembles a single step in a standard neural network convolution, it functions differently. In a typical convolution, the data is processed using trainable filters. Here, however, the correlation layer compares data with other data, eliminating the need for learnable weights.

\subsection{Moving object segmentation decoder}
We concatenate the correlation map and BEV feature map of the current frame, thereby providing the BEV features of the current as a context to the motion cues from the correlation map, and the final stage of the decoder is the linear projection to reduce the number of filters, which consists of a $3\times3$  convolutional layer followed by a $1\times1$ convolutional layer to achieve the final output BEV segmentation map of moving vehicles. 

\section{\uppercase{Experimentation Details}}
To evaluate our model, we conduct experiments with different sensor combinations and different fusion methods on the publicly available nuScenes dataset \cite{nuscenes}.

\subsection{Dataset} The nuScenes dataset contains a rich collection of point cloud data and image data from 1,000 scenes, each spanning a duration of 20 seconds collected over a wide range of weather and time-of-day conditions. The data acquisition vehicle is equipped with 6 cameras, 5 radar sensors, and a 360-degree, 32-beam LiDAR scanner. We use the official nuScenes training/validation split, which contains 28,130 samples in the training set and 6,019 samples in the validation set.


\subsection{Setup}

In our baseline model, we use ResNet-101 \cite{he2016deep} for the image backbone. We downsample all input images to a resolution of $224 \times 400$. For the 2D-3D transformation, we use the same lifting strategy as described in section \ref{feature_extraction}. In the LiDAR path, we voxelize point clouds and create binary occupancy 3D grids. In the radar path, we also apply the rasterized radar BEV feature map as described in section \ref{feature_extraction}.
\vspace{-5mm}

\begin{center}
    \begin{table*}[ht]
    \captionsetup{font=small}
    \centering
    \caption{MOS with different sensor setups and fusion strategies. `C', `R', and `L' represent camera, radar, and LiDAR, $\uparrow$ indicates that a higher value is better. $\otimes$ denotes multimodal deformable cross attention, $\oplus$ denotes concatenation. * denotes our baseline model for experiments with the MDCA fusion method.}
    \scalebox{1.0}{
        \begin{tabular}{ p{5.9cm}  p{1.0cm} p{2.3cm}  p{1.8cm} p{1.0cm} p{0.8cm}}
         \hline
         Method                      &Modality& Image  backbone  & Fusion method          &  Precision (\%)$\uparrow$ &  mIoU (\%)$\uparrow$\\ [0.5ex] 
         \hline 
         MoSeg\cite{sigatapu2023bev} &    C   & EfficientNet-b0 &  -                     &   -        & 26.0 \\  
         SimpleBEV\_Motion\cite{harley2022simple}                   &    C   & ResNet-101      &  -                     &    49.43   & 34.04\\ 
         SimpleBEV\_Motion\cite{harley2022simple}                   &   C+R  & ResNet-101      & C $\oplus$ R             &    66.65   & 51.52\\ 
         SimpleBEV\_Motion\cite{harley2022simple}                   &   C+L  & ResNet-101      & C $\oplus$ L             &    67.44   & 50.27\\ 
         SimpleBEV\_Motion\cite{harley2022simple}                   &  C+R+L & ResNet-50       & C $\oplus$ R $\oplus$ L    &    73.03   & 59.86 \\
         SimpleBEV\_Motion\cite{harley2022simple}                   &  C+R+L & EfficientNet-b4 & C $\oplus$ R $\oplus$ L    &    73.25   & 59.90 \\
         SimpleBEV\_Motion\cite{harley2022simple}\text{*}           &  C+R+L & ResNet-101      & C $\oplus$ R $\oplus$ L    &    73.79   & 60.24 \\
         \hline
         BEVMOSNet                   &  C+R+L & ResNet-101      & C $\otimes$ (L $\oplus$ R) &    73.22   & 61.82\\
         BEVMOSNet                   &  C+R+L & ResNet-101      & (C $\otimes$ L) $\oplus$ R &    74.93   & 60.91\\
         BEVMOSNet                   &  C+R+L & EfficientNet-b4 & (C $\otimes$ R) $\oplus$ L &    75.05   & 62.22\\
         BEVMOSNet (ours)             &  C+R+L & ResNet-101      & (C $\otimes$ R) $\oplus$ L &    \textbf{75.35}   & \textbf{62.59}\\
         \hline 
        \end{tabular}
        \label{table:baseline_result}
    }
    \end{table*}
    \vspace{-2mm}
\end{center}

We use the setup described in SimpleBEV \cite{harley2022simple} as a baseline for our experiments. We use a $100m \times 100m$ region around the ego-vehicle (+/- 50m in front of and behind, +/- 50m left and right of the ego-vehicle) with a grid cell size of 50cm. This results in a 2D BEV grid map with a shape of $200 \times $200. Along the vertical axis, we set the range to 10 m and discretize at a resolution of 8. The 3D grid volume then is shaped as $200 \times 8 \times 200$ ($X \times Y \times Z$). We orient this 3D grid according to the reference camera. To evaluate our predicted segmentation output, we use the Intersection-over-Unition (IoU) and pixel precision metrics. IoU is the score between the prediction and the ground truth (GT) of the moving vehicle in the current frame. Precision score is the number of true positive pixels divided by the number of all positive pixels. Since these GTs are not available in the nuScenes dataset, we follow BEV-MoSeg \cite{sigatapu2023bev} to generate them. First, we filter 3D bounding boxes for vehicles within our defined grid area. Next, we utilize the 'vehicle.moving' attribute on the filtered bounding boxes to identify vehicles that are in motion. At the end, we project the filtered 3D bounding boxes into the BEV space to generate binary masks. We train our baseline model with the Adam optimizer, a learning rate of 3e-4, and a weight decay of 1e-7 using 4 A100 GPUs. For all experiments, we use a batch size of 40 and train for 75,000 iterations. 
We use the standard binary cross-entropy loss to supervise the moving vehicle segmentation:
\begin{equation}
\mathcal{L}_{BCE} = \frac{-1}{N}\sum_{i=1}^{N} y_i\log(p_i) + (1-y_i)\log(1-p_i)
\end{equation}
where $p_i$ denotes the prediction at pixel $i \in [1, N]$, and $y_i \in \{0, 1 \}$ denotes the binary ground truth label at pixel $i$, which specifies whether the pixel $i$ belongs to the vehicle class.



\subsection{Baseline experiments} \label{Baseline_experiment}
Due to the limited number of state-of-the-art baseline models for moving object segmentation in BEV, we extended SimpleBEV \cite{harley2022simple} for the moving vehicle segmentation task (SimpleBEV\_Motion) as the second baseline model besides BEV-MoSeg \cite{sigatapu2023bev} to investigate the impact of each sensor modality on the MOS task. In the fusion module, we use the simple concatenation fusion method. We start with experiments for the camera-only model. Next, we train the model using two sensor modalities, e.g., cameras with radar and cameras with LiDAR. Finally, we train our model with all sensor data (camera, radar, LiDAR). Table \ref{table:baseline_result} shows experiment results with the SimpleBEV\_Motion model. Compared with BEV-MoSeg \cite{sigatapu2023bev}, the SimpleBEV\_Motion model outperforms BEV-MoSeg \cite{sigatapu2023bev} with all sensor configurations. Even in the camera-only scenario, the model uses a simple lifting strategy without any learnable parameters and achieves an 8.04\% improvement. The SimpleBEV\_Motion model achieves state-of-the-art results in camera + radar + LiDAR fusion scenarios for the moving vehicle segmentation task. 

\subsection{Fusion experiments}
Based on the baseline experiments in \ref{Baseline_experiment}, we selected the best candidate model as SimpleBEV-Motion (C+L+R) for further experiments with the MDCA fusion method. To evaluate the MDCA for the moving object segmentation task, we apply MDCA with 3 different fusion strategies: 1. Camera-radar fusion using DCA, then the fused feature map is concatenated with the LiDAR BEV feature map; 2. Camera-LiDAR fusion using DCA, then the fused feature map is concatenated with the radar BEV feature map; 3. Radar and LiDAR feature maps are concatenated, then fused with the camera feature map using DCA. Table \ref{table:baseline_result} shows our experiment results with these configurations. We confirm that the MDCA strategy can boost the performance of moving object segmentation tasks in all fusion configurations compared to the baseline. We also observe that by applying MDCA for camera and radar features, we can utilize the dynamic information inherited in radar data to help the model focus on more useful camera features, which helps to improve the model performance. 

\begin{figure*}[h!]
\captionsetup{font=small}
    \centering
    \includegraphics[width=\textwidth]{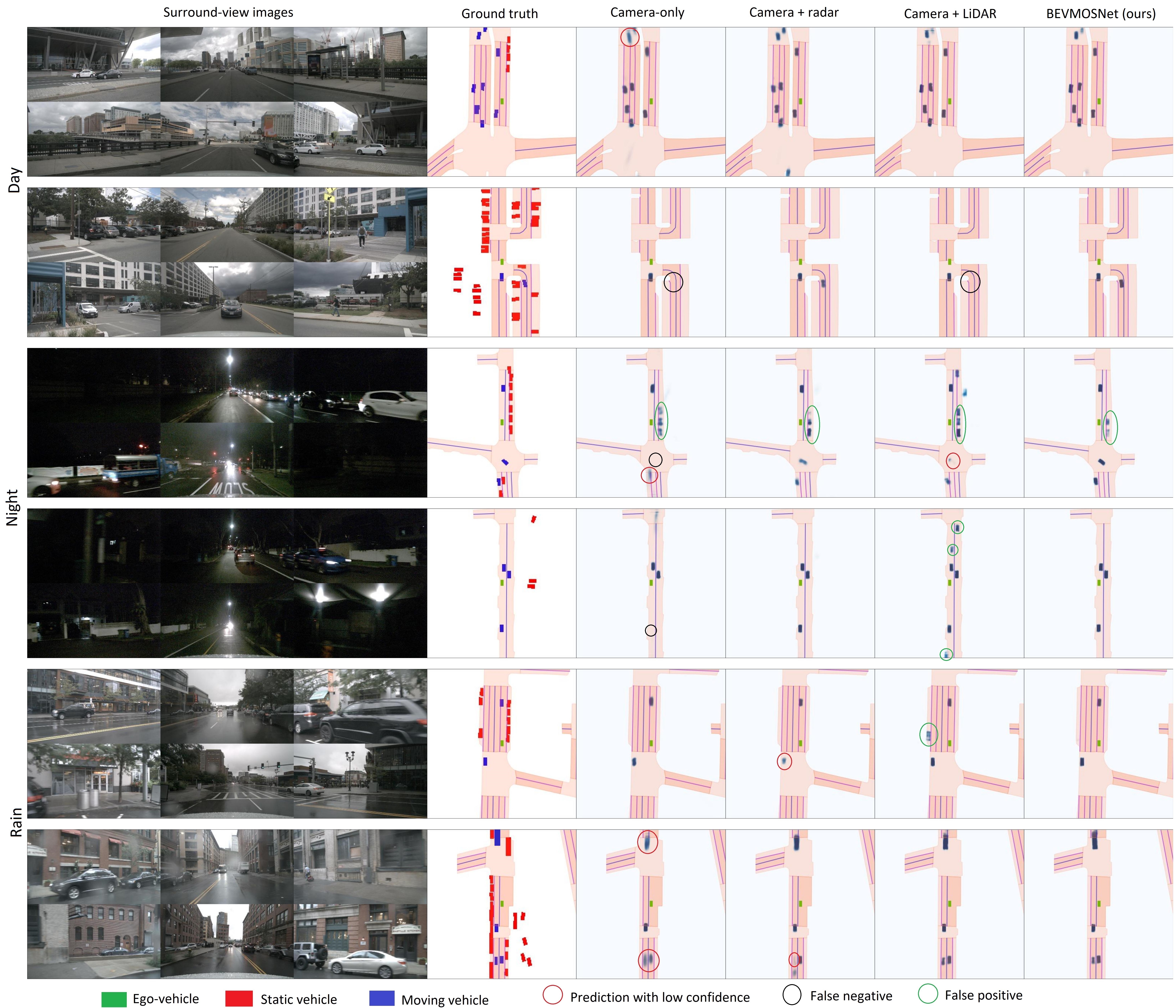}
    \caption{Qualitative results on MOS in various weather conditions. The camera-only model predicts distant moving objects with lower confidence (blurred region, marked with red circles). It also fails to segment occluded moving objects, or when operating in low light conditions, such as at night (regions marked with black circles). Generally, LiDAR helps to locate object positions and estimate object orientation accurately; radar improves the segmentation of distant objects. By combining camera, LiDAR, and radar we can leverage the advantages of each modality to build a robust model, which reduces false positive predictions (marked with green circles). }
    \label{fig:qualitative results}
\end{figure*}

\subsection{Ablation Study}
For each ablation experiment, we only train models for 50,000 iterations for faster convergence. Table \ref{table:Iou over distance} shows the IoU of segmented moving objects over different distances. The performance of the camera-only model drops significantly when objects are far from the ego-vehicle. With LiDAR data, the camera + LiDAR model performs better on far objects. Due to the sparsity and noise of radar data, the camera-radar model performs slightly worse at closer distances compared to camera-LiDAR, but it improves significantly at long distances.

In Table \ref{table:Aggregated LiDAR and Radar Frames}, we show the result of experiments with different numbers of aggregated LiDAR and radar sweeps. We observe that aggregation of sweeps helps to improve the model performance, and the baseline model achieves the best performance with an aggregation of five sweeps; therefore, we conduct all experiments in Table \ref{table:baseline_result} with the five-sweep aggregation. It has been observed that when we consider more than five sweeps, then we see performance degradation. Although the corresponding results are not reported, this is a critical finding, as we observed accumulating more sweeps shows almost zero overlap of the moving object across frames. This phenomenon can be attributed to the predominance of urban scenarios in nuScenes dataset, where dynamic objects such as vehicles traverse considerable distances over time. In Table \ref{table:mIoU_vs_weather} we show the model performance in daytime, nighttime and rain driving conditions where the performance in rain scenes is significantly improved by using multimodal sensor fusion. The performance gap of the camera-only model between rain and daytime conditions is 6.05\%. By adding radar data, this gap is shrunk to 2.81\%. The camera-LiDAR-radar model closes this gap and increases the performance in rain conditions by 0.84\% compared to the performance in daytime conditions. This confirms that radar and LiDAR are useful for perception in adverse weather conditions. Generally, adding radar and LiDAR helps improve the performance of the camera-only model in all driving conditions.

\begin{center}
    \begin{table}[ht]
    \captionsetup{font=small}
    \centering
    \caption{Ablation study on the usage of unimodal vs. multimodal sensor using mIoU metric with respect to distances for moving object segmentation task.}
        \begin{tabular}{ p{2.0cm} p{1.0cm} p{1.2cm} p{1.2cm}}
         \hline
                          &    0-20m        &     20-35m     & 35-50m\\ [0.5ex] 
         \hline 
         C                &    50.14        &     35.91      &  19.11          \\  
         C+R              &    61.10         &     55.57      &  41.31          \\ 
         C+L              &    63.73        &     55.11      &  35.96          \\
         BEVMOSNet &  \textbf{72.68} & \textbf{67.04} & \textbf{51.21} \\ 
         \hline
        \end{tabular}
    \label{table:Iou over distance}
    \end{table}
    \vspace{-7mm}
\end{center}

\begin{center}
    \begin{table}[ht]
    \captionsetup{font=small}
    \centering
    \caption{Exploring the influence of varying numbers of aggregated LiDAR and radar sweeps with camera data. The best sweep aggregation was chosen for our experiments in Table \ref{table:baseline_result}.}
        \begin{tabular}{ p{1.2cm} p{0.9cm} p{0.8cm} p{1.3cm} p{1.2cm}}
         \hline
         & C+L & C+R & C+R+L baseline & BEV- MOSNet\\ [0.5ex] 
         \hline 
          1 frame   & 46.73 & 50.13 & 58.69 & - \\  
          3 frames  & 48.20 & 51.01 & 59.68 & - \\ 
          5 frames  & 50.27 & 51.52 & 60.24 & \textbf{62.59}\\ 
         \hline
        \end{tabular}
    \label{table:Aggregated LiDAR and Radar Frames}
    \end{table}
    \vspace{-7mm}
\end{center}

\begin{center}
    \begin{table}[h!]
    \captionsetup{font=small}
    \centering
    \caption{Performance analysis of BEVMOSNet and comparison with other sensor proposals in adverse weather scenarios and low illumination conditions using mIoU.}
        \begin{tabular}{ p{1.0cm} p{1.0cm} p{0.9cm} p{0.9cm} p{1.3cm}}
         \hline
                  & C &    C+L  &   C+R   & BEV- MOSNet\\ [0.5ex] 
         \hline 
          Rain    &    28.75    &  50.88  &  49.04  & \textbf{63.41} \\  
          Day     &    34.80    &  49.67  &  51.85  & \textbf{62.57} \\
          Night   &    35.10    &  52.76  &  51.44  & \textbf{59.64} \\
         \hline
        \end{tabular}
    \label{table:mIoU_vs_weather}
    \end{table}
    \vspace{-5mm}
\end{center}

\subsection{Qualitative results}
Figure \ref{fig:qualitative results} shows qualitative results of moving vehicle segmentation. We observe that the prediction confidence of the camera-only model is lower, particularly for objects far from ego vehicles. The camera-LiDAR model predicts more precise object locations and also increases prediction confidence. The camera-radar model helps greatly predict objects at far distances. Besides that, radar sensors also provide information about dynamic objects through velocity attributes, which greatly improves the prediction of occluded moving objects. On the other hand, due to the sparse and noisy nature of radar data, the camera-radar model produces more noisy predictions. By combining all three sensor modalities, we achieve a more robust model, which compensates for the weaknesses of each sensor type and increases the overall segmentation performance. 

\section{\uppercase{Conclusion}}

In this work, we introduce a novel multi-sensor, multi-camera architecture for motion understanding in BEV, achieving a 62.59\% IoU score on the nuScenes moving object detection dataset. Our investigation includes extensive experiments aimed at assessing the impact of each sensor modality on overall performance during the feature fusion stage and optimal configuration for sensor fusion. Additionally, we integrate deformable cross-attention to improve the extraction of robust camera features, leveraging the complementary information from LiDAR and radar modalities. Due to the limited availability of moving object labels within nuScenes, which are currently restricted to the vehicle class, our experimental validation solely focuses on this category. However, it is possible to boost the performance further and extend the motion detection task to more classes, such as bicyclists and pedestrians with the label availability. We leave this work to future research.

\bibliographystyle{apalike}
{\small
\bibliography{example}}



\end{document}